\documentclass[runningheads]{llncs}

\usepackage[T1]{fontenc}
\usepackage{graphicx}
\usepackage{booktabs}
\usepackage{amsmath}
\usepackage{amssymb}
\usepackage{multirow}
\usepackage{microtype}
\usepackage{xspace}

\newcommand{\eg}{\textit{e.g.,}\xspace}

\newcommand{\etal}{\textit{et al.}\xspace}

\begin{document}

\title{SGMCE: Segment-Grounded Morphological Concept Explanation
for Malaria Parasite Species Identification in Thick Blood Smears}

\titlerunning{SGMCE for Malaria Parasite Species Identification}


\author{Ahmed Tahiru Issah\inst{1} \and
Charles B. Delahunt\inst{2} \and
Carine Mukamakuza\inst{1}\thanks{Corresponding author.}}
\authorrunning{A.T. Issah et al.}
%
\institute{Carnegie Mellon University, Kigali, Rwanda\\
\email{aissah@alumni.cmu.edu, cmukamak@andrew.cmu.edu} \and
University of Washington, Seattle, Washington\\
\email{delahunt@uw.edu}}


\maketitle

\begin{abstract}
Malaria diagnosis in endemic regions depends on species-level identification of \textit{Plasmodium} parasites in thick blood smears, but deep learning detectors classify detections without providing morphological evidence for their predictions, limiting the ability of microscopists to audit those predictions at the case level. We present SGMCE (Segment-Grounded Morphological Concept Explanation), a post-hoc explanation framework that requires no additional training, no morphological annotations, and no labelled explanation data, yet produces per-detection natural-language explanations anchored in thick-smear morphology. For each detection, SGMCE extracts mask-guided crop thumbnails, computes fourteen handcrafted computer-vision morphological features (shape, colour, chromatin, haemozoin pigment) using adaptive within-mask thresholds, and queries GPT-4o with both visual evidence and computed measurements, conditioned on a thick-smear-specific knowledge base compiled from the World Health Organization bench aids. The primary output is a structured explanation identifying which morphological features support the detected species and why the competing species are excluded. Explanations are validated by four automatic metrics: Knowledge-Base Consistency (KBC), CV-Claim Faithfulness (CCF), Discriminativeness Score (DS), and LLM-as-Judge (LLMj). A sentence-level semantic scoring rule with species-aware negation filtering resolves the vocabulary mismatch between clinical prose and knowledge-base terms. Across 737 detections from 139 thick-smear images spanning four \textit{Plasmodium} species and white blood cells, parasite-class mean KBC is 0.91, mean DS is 0.99, and mean CCF is 0.97, while a per-rule CCF breakdown confirms that the CV-grounded claims made by the vision-language model are consistent with the measurements they cite.

\end{abstract}

\section{Introduction}
\label{sec:intro}

Malaria, caused by \textit{Plasmodium} parasites transmitted by female
\textit{Anopheles} mosquitoes, remains one of the most significant preventable causes of mortality worldwide, with roughly 249\,million cases and 608{,}000 deaths estimated by the World Health Organization in 2022, and more than 90\,\% of global mortality concentrated in Sub-Saharan Africa~\cite{who2023}. Species-level identification is clinically important because treatment regimens differ across species: \textit{P.\,falciparum} (Pf) requires artemisinin-based combination therapy and, if untreated, can progress to fatal cerebral malaria within hours; \textit{P.\,vivax} (Pv) and \textit{P.\,ovale} (Po) additionally require primaquine to clear dormant liver hypnozoites and prevent relapse; and \textit{P.\,malariae} (Pm) typically responds to chloroquine alone~\cite{who_treatment_2015,cdc_treatment}.
Thick blood smear (thick-film) microscopy remains the diagnostic gold
standard in endemic regions because of its high sensitivity at low
parasitaemia~\cite{poostchi_image_2018}, but it requires expert microscopists, is time-consuming, and suffers from inter-observer
variability~\cite{delahunt_metrics_2024}.

Deep learning object detectors and segmentation models have demonstrated strong object-level accuracy on parasite detection in blood smears~\cite{ramos-briceno_deep_2025,boit_efficient_2024,delahunt_fully-automated_2019}. Object-level detection, locating and classifying individual parasites in a full field of view, is a key step in patient-level diagnosis, which requires aggregating object-level counts across a slide to estimate parasitaemia and infer infection status~\cite{delahunt_metrics_2024,delahunt_fully-automated_2019}.
A separate requisite for clinical utility, orthogonal to raw accuracy, is to ensure user trust by explaining \emph{why} a detection is assigned to a particular species. This matters for microscopist-in-the-loop review, where a diagnostician auditing an automated prediction needs to know the logic behind the prediction (\eg whether the chromatin dots are single or double), and why the observed morphology excludes competing species.

Existing explainability methods such as Grad-CAM~\cite{selvaraju_grad-cam_2017},
LIME~\cite{ribeiro_why_2016}, and SHAP~\cite{lundberg_unified_2017} all produce pixel-level saliency maps or attribution scores. These identify regions of interest but not the
morphological reasoning behind a prediction. A saliency heatmap cannot indicate whether a bright region is a chromatin dot or a staining artefact, and it produces no differential argument against competing species. Concept-based approaches such as TCAV~\cite{kim_tcav_2018} explain model behaviour in terms of human-defined concepts, but require trained probes and do not generate per-detection free-form explanations. Closest to our setting, MorphXAI~\cite{yousaf_morphxai_2026} couples parasite detection with fine-grained morphological analysis on \textit{Leishmania} and \textit{Trypanosoma} species and fills in a templated text, demonstrating that structured morphological attributes (shape, dot count, developmental stage) can be integrated into detection pipelines to produce explanations closer to the evidence clinicians rely on than pure saliency. Our work starts from the same premise - that morphology, rather than pixel intensity, is the right vocabulary for parasitology explanations, and extends it to \textit{Plasmodium} speciation
in thick films, replacing fixed template sentences with free-form reasoning from a vision-language model in a post-hoc, knowledge-base-grounded pipeline. We use GPT-4o~\cite{openai_gpt-4o_2024} as the reasoning backbone. Recent work~\cite{moor_foundation_2023} reports strong zero-shot performance of modern vision-language models on medical image understanding tasks when they are appropriately prompted with structured domain knowledge, which suits the per-detection explanation setting we target.

A practical caveat is that reliable non-falciparum speciation on thick film is acknowledged to be difficult even for expert microscopists. The standard WHO bench aids note that \emph{Pv}, \emph{Po}, and \emph{Pm} are often indistinguishable on thick film and that confirmatory speciation is usually performed on a thin film~\cite{who_benchaids}. The YOLOv12n detector used in this study was trained on thick-smear images with clinical species labels, so its predictions reflect patterns learnable from thick film under the labelling conventions of that laboratory. SGMCE explains those predictions using thick-smear-appropriate
morphological evidence. It does not claim to resolve cases that would be ambiguous even under expert review.

We introduce \textbf{SGMCE} (Segment-Grounded Morphological Concept Explanation), a hybrid computer-vision (CV) plus vision-language model (VLM) post-hoc framework. For each detected parasite, SGMCE \emph{(i)} identifies species-discriminating morphological features supporting
the classification; \emph{(ii)} provides differential reasoning explicitly rejecting competing
species; \emph{(iii)} describes the evidence in clinical vocabulary; and \emph{(iv)} self-validates through four metrics.

\paragraph{Contributions.}
\begin{enumerate}
  \item A post-hoc LLM-based explanation pipeline for malaria speciation that requires         no model retraining, no morphological training annotations, and is applicable to any segmentation use case where the model outputs instance masks.
  \item A thick-smear-specific morphological knowledge base encoding per-species hallmarks and pairwise differentiators, compiled from the WHO bench aids for the diagnosis of malaria infections~\cite{who_benchaids}.
  \item A handcrafted CV morphological feature extractor computing fourteen clinically grounded measurements from mask-guided crops using adaptive within-mask thresholds suited to variable thick-smear staining.
  \item Four automatic explanation-validation metrics, including a sentence-level semantic embedding scoring rule with species-aware negation filtering that resolves vocabulary mismatch between clinical prose and knowledge-base keyword strings.
  \item Empirical evaluation on 737 detections from 139 held-out thick-smear images across four \textit{Plasmodium} species, showing parasite-class mean KBC of 0.91, DS of 0.99, and CCF of 0.97, with a per-rule CCF breakdown that supports the faithfulness of the VLM's visual claims.
\end{enumerate}

\section{SGMCE Framework}
\label{sec:framework}

\subsection{Pipeline Overview}
\label{sec:overview}

\begin{figure}[tb]
\centering
\includegraphics[width=\textwidth]{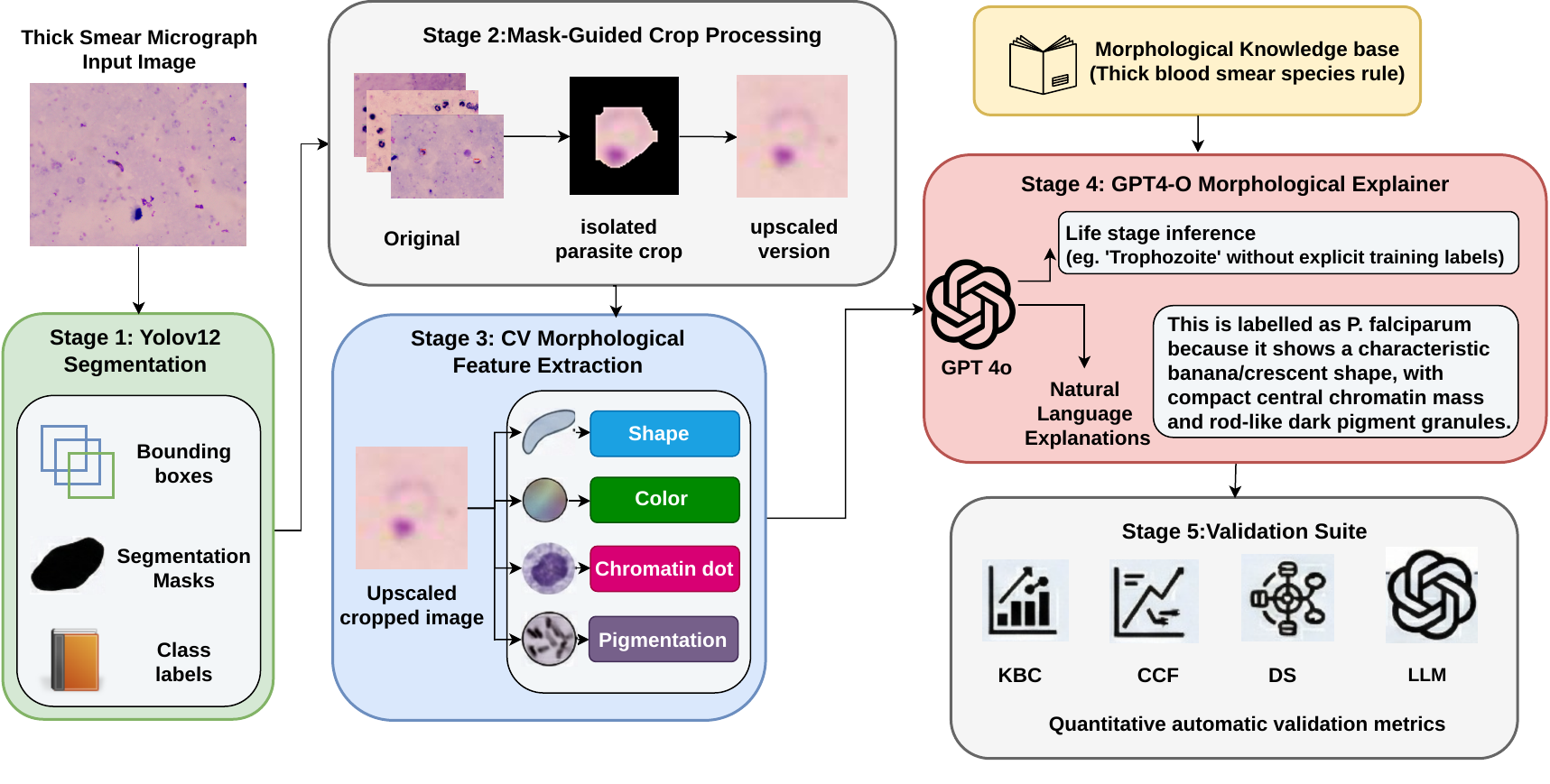}
\caption{SGMCE pipeline for thick-film malaria explanation. Each detected object passes through five stages. GPT-4o receives two upscaled crop images (visual path) and a structured text rendering of the fourteen CV measurements (measurement path), conditioned on a thick-smear-specific morphological knowledge base (KB) compiled from WHO guidelines.}
\label{fig:pipeline}
\end{figure}

Figure~\ref{fig:pipeline} illustrates the five-stage SGMCE pipeline. Given a trained segmentation model (here, a Yolov12) and a thick-smear image, SGMCE produces for each detected object: two mask-guided cropped thumbnails, fourteen CV morphological measurements, a species-discriminating natural-language explanation with differential reasoning, and four automatic validation scores. Stages~1--2 (detection and crop preprocessing) are standard components adapted for thick-smear constraints. Stages~3--5 (handcrafted morphological feature extraction, knowledge-base-grounded GPT-4o reasoning, and the automatic validation suite) are the novel contributions of this work. All stages are post-hoc: no model weights are read or modified at any stage.

\subsection{YOLOv12n Segmentation Detector}
\label{sec:yolo}

We use a YOLOv12n-seg model~\cite{tian_yolov12_2025} trained on approximately 14{,}000 thick-smear images (after class-balancing augmentation) from a national biomedical centre in Sub-Saharan Africa, classifying five categories: \emph{Pf}, \emph{Pv}, \emph{Po}, \emph{Pm}, and white blood cells (WBCs). Training runs for 70 epochs at resolution $2048{\times}2048$ with SGD (momentum 0.937, cosine learning-rate schedule). At inference, we apply a confidence threshold of 0.25. Each detection yields a bounding box $(x_1,y_1,x_2,y_2)$, an instance mask $M \in \{0,1\}^{H \times W}$, a class label $c$, and a confidence $\hat{p}$. All detections above the confidence threshold - including any debris, staining artefacts, or distractor objects that the detector classifies as parasites - are passed through the subsequent SGMCE stages unchanged; low-confidence detections are flagged diagnostically but not suppressed
(Sect.~\ref{sec:fp}).

\subsection{Mask-Guided Crop Preprocessing}
\label{sec:crops}

Raw detections span only 10--60\,px in our acquisition setting, which is too small for reliable visual analysis by a VLM. The \emph{CropPreprocessor} extracts two complementary thumbnails per detection (examples in Figure~\ref{fig:pipeline}, stage 2 box).

\noindent\textbf{1. Masked crop.} The bounding-box region is extracted and pixels outside mask $M$ are replaced with the image's estimated background colour (sampled from the image corners, yielding the characteristic pinkish Giemsa-stained smear background), isolating the parasite from surrounding lysed red-blood-cell debris.

\noindent\textbf{2. Context crop.} The bounding box is symmetrically padded by 50\,\% of its width and height before extraction, preserving surrounding context (nearby parasites, co-detected WBCs, pigment distribution).

Both crops are upscaled to a minimum dimension of 200\,px (4--12$\times$ magnification) via bicubic interpolation, which preserves edge sharpness at these magnification factors and is important for chromatin-dot boundary delineation. Crops are encoded as JPEG images for transmission to GPT-4o.

\subsection{Handcrafted Morphological Feature Extractor}
\label{sec:cv}

The \emph{CVFeatureExtractor} computes fourteen handcrafted morphological
measurements from the masked crop (Table~\ref{tab:features}), grounded in clinical practice \cite{who_benchaids}. Thresholds are adaptive, derived from within-mask pixel statistics, to accommodate the variable staining intensity typical of field-prepared thick smears.

\noindent\textbf{Shape analysis.}
Shape is computed from the binary mask contour. Let $A$ denote pixel area and $P$ the contour perimeter. \emph{Circularity} is $C = 4\pi A / P^2 \in [0,1]$, where $C=1$ is a perfect circle and $C>0.65$ is characteristic of ring-stage parasites and WBCs. Fitting the minimum-area enclosing ellipse yields major/minor semi-axes $a,b$ with aspect ratio $\mathrm{AR}=a/b$ and eccentricity $e = \sqrt{1-(b/a)^2}$; $\mathrm{AR}>2.5$ and $e>0.90$ flag the characteristic falciform \emph{Pf} gametocyte. \emph{Solidity} $S = A / A_{\mathrm{hull}}$ (where $A_{\mathrm{hull}}$ is the convex-hull area) captures amoeboid cytoplasmic extensions typical of \emph{Pv} growing trophozoites when $S<0.82$.

\noindent\textbf{Colour analysis.}
The masked crop is converted from BGR to CIE~LAB and HSV colour spaces, and statistics are computed over mask pixels only. \emph{Blue-purple fraction} $\mathrm{BP}$ is the proportion of mask pixels with OpenCV HSV hue $h_i \in [100, 150]$ (equivalent to $[200^{\circ}, 300^{\circ}]$ in standard convention), capturing Giemsa nucleic-acid staining. \emph{Dark fraction} $\mathrm{DF}$ is the proportion of mask pixels with HSV value $V_i < 80$, indicating haemozoin or densely packed chromatin.

\noindent\textbf{Chromatin detection.}
Chromatin appears as small bright reddish dots under Giemsa staining and is most prominent on the CIE~LAB $a^*$ axis -- the red-to-green opponent channel, where positive values indicate red. Let $\mu_a$ and $\sigma_a$ be the within-mask mean and standard deviation of $a^*$. Candidate chromatin pixels satisfy
\begin{equation}
  a^*_i > \mu_a + \max(1.2\,\sigma_a,\; 8.0),
  \label{eq:chrom}
\end{equation}
where the absolute offset 8.0 ensures a meaningful threshold when $\sigma_a$ is low. Connected components whose area is between 0.5\,\% and 15\,\% of the mask area yield the chromatin count $N_c$. The \emph{double-chromatin flag} ($N_c \geq 2$) is characteristic of Pf ring-stage parasites.

\noindent\textbf{Pigment (haemozoin) detection.}
Haemozoin appears as dark brownish-black granules with species-specific colour, count, and distribution. Detection uses the CIE $L^*$ (luminance) channel; pixels satisfying $L^*_i < \mu_L - \max(\sigma_L,\; 10.0)$ are haemozoin candidates. Components whose area is at least 0.5\,\% of the mask area contribute to the granule count $N_p$; the normalised mean distance of granule centroids from the mask centroid classifies the distribution as \emph{central} ($<35\,\%$ of the maximum radius), \emph{peripheral} ($>60\,\%$), or \emph{scattered}.

\noindent\textbf{Size reference via co-detected WBC.}
WBCs are reliably abundant in thick films, and lie in a well-characterised size range, typically 10--14\,\textmu, which projects to roughly 60--120\,px in our $2048{\times}2048$ acquisitions, whereas ring-stage parasites span 10--30\,px. When a WBC is co-detected in the same image, the parasite-to-WBC area ratio is reported per detection as \texttt{relative\_size\_vs\_wbc}. Typical values are 0.03--0.15 for ring-stage parasites, 0.2--0.6 for larger trophozoites or schizonts, and close to 1.0 for co-detected WBCs themselves. This ratio gives GPT-4o a grounded size cue that does not rely on intact red blood cells, which are absent in thick films due to lysing.

\begin{table}[tb]
\caption{The fourteen handcrafted CV morphological features computed by
\texttt{CVFeatureExtractor}. All threshold-derived flags are adaptive relative
to within-mask pixel statistics.}
\label{tab:features}
\centering
\small
\begin{tabular}{p{3.3cm}lp{6.0cm}}
\toprule
Feature & Type & Morphological interpretation \\
\midrule
\texttt{circularity}            & Float $[0,1]$ & Compactness; high ($>0.65$) for rings or WBCs \\
\texttt{aspect\_ratio}          & Float         & Elongation; $>2.5$ flags Pf gametocyte \\
\texttt{eccentricity}           & Float $[0,1]$ & Ellipse axis ratio; high for crescent shape \\
\texttt{solidity}               & Float $[0,1]$ & Convexity; $<0.82$ flags amoeboid Pv \\
\texttt{is\_round}              & Bool          & $C > 0.65$ \\
\texttt{is\_elongated}          & Bool          & AR $> 2.0$ \\
\texttt{is\_amoeboid}           & Bool          & $S < 0.82$ \\
\texttt{blue\_purple\_fraction} & Float         & Giemsa nucleic-acid stain uptake \\
\texttt{dark\_fraction}         & Float         & Dark pixel fraction (haemozoin / chromatin) \\
\texttt{chromatin\_count}       & Int           & Chromatin dots $N_c$ \\
\texttt{has\_double\_chromatin} & Bool          & $N_c \geq 2$; key Pf ring indicator \\
\texttt{pigment\_count}         & Int           & Haemozoin granules $N_p$ \\
\texttt{pigment\_distribution}  & String        & Central / peripheral / scattered \\
\texttt{relative\_size\_vs\_wbc}& Float         & Parasite-to-WBC area ratio (when WBC available) \\
\bottomrule
\end{tabular}
\end{table}

\subsection{Thick-Smear Morphological Knowledge Base}
\label{sec:kb}

The knowledge base (KB) is compiled from the WHO bench aids for the diagnosis of malaria infections~\cite{who_benchaids}, the standard thick-smear parasitology reference used for laboratory training in endemic regions. The KB is deliberately thick-film-specific: features that depend on intact red blood cells are excluded, because RBC enlargement (Pv) and Ziemann's stippling (Pm) are not visible after RBC lysis. Schüffner's dots, visible as discrete pink stipples on thin films, appear on thick films as a diffuse pinkish cloud around the parasite; the KB describes this thick-film appearance explicitly to avoid priming the VLM with thin-film terminology.

The KB contains one entry per class (\emph{Pf, Pv, Po, Pm, WBC}). Each entry records \texttt{thick\_film\_notes} (general appearance); \texttt{life\_stages} (per-stage feature lists for ring, growing trophozoite, mature trophozoite, schizont, and gametocyte); \texttt{key\_differentiators} (three to five most discriminative thick-film features); \texttt{pigment} (haemozoin colour, distribution, count); and \texttt{differential\_vs} (pairwise feature lists against each alternative species). Serialised as structured plain text, the KB is embedded verbatim in the GPT-4o system prompt for every API call, ensuring consistent grounding across all detections. The full system prompt (KB plus thick-smear context preamble, reasoning guidance, and output JSON schema) occupies approximately 1{,}500 tokens and contains no embedded images.


\subsection{GPT-4o Morphological Explainer}
\label{sec:explainer}

The \emph{MorphologicalExplainer} queries GPT-4o through the OpenAI Chat
Completions API. Each call comprises a structured system prompt, a
per-detection user prompt, and two base64-encoded JPEG images
(masked and context crops).

\noindent\textbf{System prompt.} Four parts:
\emph{(i)} a thick-smear context preamble that directs GPT-4o away from
thin-smear reasoning patterns;
\emph{(ii)} the full KB from Sect.~\ref{sec:kb};
\emph{(iii)} reasoning guidance that prioritises species-discriminating
morphological evidence and that requires an explicit rejection of each
alternative species before reporting uncertainty;
\emph{(iv)} the output JSON schema, specifying every output field's name, type,
and content guidance, with \texttt{supporting\_features} and
\texttt{differential\_reasoning} designated the primary outputs.
A separate \texttt{WBC\_SYSTEM\_PROMPT} is used for WBC detections. It focuses
on nuclear architecture (neutrophil, lymphocyte, monocyte, eosinophil) and
explicitly prohibits parasite-feature discussion, preventing the VLM from
over-reading morphology on non-parasite objects.

\noindent\textbf{User prompt.} Three items are passed per detection, in
addition to the two images:
\emph{(i)}~the YOLO class label $c$ and confidence $\hat{p}$, presented as a
prior belief rather than ground truth;
\emph{(ii)}~the fourteen CV measurements rendered as a structured text block,
pairing each numeric value with a human-readable interpretive label so that
GPT-4o can treat it as a natural-language cue rather than a raw number, \eg:
\begin{quote}\small\ttfamily
Circularity: 0.82 (round), AR: 1.04, Chromatin dots: 2 (DOUBLE --\\
notable in Pf ring stage), Pigment granules: 1 (central),\\
Blue-purple fraction: 0.47, Size vs WBC: 0.06 (parasite is very\\
small relative to co-detected WBC).
\end{quote}
\emph{(iii)}~a per-species hallmark checklist, asking GPT-4o to respond
\texttt{observed}, \texttt{not\_observed}, or \texttt{uncertain} for each
thick-film hallmark of the detected class (\eg for Pf: ``double chromatin
dot'', ``small delicate ring'', ``falciform gametocyte'').

\noindent\textbf{Structured output.} A JSON object with primary fields
\texttt{supporting\_features}
(feature--observation--significance triples grounding specific claims in
visual evidence); \texttt{differential\_reasoning} with explicit per-species
keys (\texttt{vs\_pf}, \texttt{vs\_pv}, \texttt{vs\_po}, \texttt{vs\_pm}),
each stating why the alternative species is excluded;
\texttt{hallmark\_checklist} (per-hallmark verdict);
\texttt{natural\_language\_explanation} (a free-form clinical description,
2--4 sentences); and \texttt{explanation\_confidence}. Secondary fields include
\texttt{estimated\_life\_stage} and \texttt{caveats}. A partial example for a
Pm schizont:
\begin{quote}\small\ttfamily
"natural\_language\_explanation":\\
\ \ "The observed morphology is consistent with Plasmodium\\
\ \ malariae, primarily due to the presence of a rosette\\
\ \ schizont formation and central pigment granule. These\\
\ \ features are pathognomonic for Pm and exclude other\\
\ \ species.",\\
"differential\_reasoning": \{\\
\ \ "vs\_pf": "Pf is excluded due to the absence of very\\
\ \ \ \ small rings and double chromatin dots, and the\\
\ \ \ \ presence of a rosette formation which is not seen\\
\ \ \ \ in Pf.",\\
\ \ "vs\_pv": "Pv is excluded because the amoeboid shape\\
\ \ \ \ is not typical for Pm, and there is no\\
\ \ \ \ golden-brown pigment.",\\
\ \ "vs\_po": "Po is excluded due to the absence of early\\
\ \ \ \ prominent stippling and the presence of a rosette\\
\ \ \ \ formation, which is not characteristic of Po." \}
\end{quote}
We use temperature $0.05$ and up to $1{,}800$ output tokens; life-stage
inference is reported as a secondary field but is not evaluated quantitatively
here, because the training labels encode species only.

\subsection{Automatic Validation Suite}
\label{sec:metrics}

Four metrics assess explanation quality without expert re-annotation and
without any manual spot-checking of individual explanations.
KBC, CCF, and DS are newly introduced in this work; LLMj adapts the
general LLM-as-judge paradigm of Zheng \etal~\cite{zheng_judging_2023} to
the per-detection setting.

\noindent\textbf{Knowledge-Base Consistency (KBC).}
KBC measures the proportion of species-appropriate thick-film morphological
concepts that are semantically present in explanation $E$. For each species we
curate a positive concept set
$\mathcal{P} = \{p_1,\ldots,p_m\}$ drawn verbatim from the species'
\texttt{key\_differentiators} (\eg for Pf: ``banana-shaped gametocyte'',
``double chromatin dot'', ``delicate ring''; for Pv: ``amoeboid cytoplasm'',
``golden-brown pigment'').
$E$ is split into sentences $\{s_1,\ldots,s_n\}$ and each sentence and concept
is independently embedded with OpenAI
\texttt{text-embedding-3-small} ($d{=}1536$). The sentence-level maximum
cosine similarity is
\begin{equation}
  f(E, p_k) = \max_{j \in [n]} \cos\!\bigl(\phi(s_j),\,\phi(p_k)\bigr),
  \label{eq:maxsim}
\end{equation}
which avoids the signal dilution that a whole-document embedding would
introduce. KBC is then
\begin{equation}
  \mathrm{KBC} \;=\; \frac{1}{m} \sum_{k=1}^{m}
    \mathbb{1}\!\bigl[f(E,p_k) \geq \theta\bigr],
  \qquad \theta = 0.40.
  \label{eq:kbc}
\end{equation}
Competing-species concepts (\eg Pv hallmarks evaluated against a Pf
explanation) are computed and logged for diagnostic inspection but are
excluded from the score, because sentence-level embeddings are negation-blind:
a sentence such as ``unlike the amoeboid morphology of \textit{P.\,vivax},
this parasite is compact'' scores high cosine similarity to the Pv amoeboid
concept even though it correctly rejects it. Penalising such sentences would
actively discourage differential reasoning, one of the primary SGMCE outputs.

\noindent\textbf{Discriminativeness Score (DS).}
DS is computed identically to KBC (Eqs.~\ref{eq:maxsim}--\ref{eq:kbc}) but
over a curated hallmark subset $\mathcal{H} \subseteq \mathcal{P}$ of
species-hallmark phrases most reliably distinguishing the detected species
from all others (\eg ``rosette schizont'' for Pm, ``banana-shaped
crescent gametocyte'' for Pf, ``amoeboid fragmented cytoplasm'' for Pv):
\begin{equation}
  \mathrm{DS} \;=\; \frac{1}{q} \sum_{k=1}^{q}
    \mathbb{1}\!\bigl[f(E,h_k) \geq \theta\bigr].
  \label{eq:ds}
\end{equation}

\noindent\textbf{CV-Claim Faithfulness (CCF).}
CCF checks whether the explanation's visual claims are consistent with the computer vision (CV) measurements. A set of seven handcrafted correspondence rules
$\mathcal{R}$ (Table~\ref{tab:ccf_rules}) maps each CV measurement to the
clinical vocabulary a microscopist would use for it. Each rule activates only
when its trigger keywords appear in the explanation; it then verifies that
the CV measurement satisfies the expected condition. To avoid the same
negation issue as in KBC, CCF applies the rules only to the
\texttt{natural\_language\_explanation} and \texttt{supporting\_features}
fields (not the \texttt{differential\_reasoning} field), and passes each
candidate clause through a species-aware negation filter that skips
clauses of the form ``unlike Pv\ldots'' or ``no amoeboid cytoplasm''.
CCF is the fraction of activated rules that agree with the corresponding CV
measurement.

\begin{table}[tb]
\caption{The seven CV-Claim Faithfulness (CCF) correspondence rules. Each
rule activates when its trigger keywords appear in the explanation; the CV
condition is then checked. Negation-scoped clauses are skipped before
activation.}
\label{tab:ccf_rules}
\centering
\small
\begin{tabular}{p{5.2cm}p{3.2cm}l}
\toprule
Trigger keywords & CV feature & Condition \\
\midrule
elongated, banana, crescent, sausage     & \texttt{aspect\_ratio}          & $> 2.0$  \\
round, circular, compact shape           & \texttt{circularity}            & $> 0.65$ \\
amoeboid, irregular, fragmented          & \texttt{solidity}               & $< 0.85$ \\
double chromatin, two dots, two chromatin & \texttt{chromatin\_count}      & $\geq 2$ \\
single chromatin, one dot, single dot    & \texttt{chromatin\_count}       & $= 1$    \\
dark pigment, hemozoin, pigment granule  & \texttt{has\_pigment}           & True     \\
blue cytoplasm, purple cytoplasm         & \texttt{blue\_purple\_fraction} & $> 0.15$ \\
\bottomrule
\end{tabular}
\end{table}

\noindent\textbf{LLM-as-Judge (LLMj).}
LLMj adapts the LLM-as-judge paradigm~\cite{zheng_judging_2023} by prompting
an independent GPT-4o instance (blind to the original system prompt) to score
factual accuracy (FA), differential specificity (SP), and clinical utility
(CU), each on $[0,1]$, yielding
$\mathrm{LLMj} = (\mathrm{FA}+\mathrm{SP}+\mathrm{CU})/3$. LLMj costs approximately \$0.01--0.03 per explanation and is reported as a complementary
diagnostic; the main validation in Sect.~\ref{sec:experiments} relies on the
three deterministic metrics KBC, CCF, and DS.

\section{Experiments and Results}
\label{sec:experiments}

\subsection{Dataset and Configuration}
\label{sec:dataset}

\noindent\textbf{Training data.}
The YOLOv12n-seg detector is trained on approximately 7{,}000 original thick
blood smear images from a national biomedical centre in Sub-Saharan Africa,
with species-specific offline augmentation and a noise-injection pass that
expands the effective training volume to roughly 14{,}000 samples. All
images are acquired under Giemsa staining at high magnification
(oil-immersion). Exact per-image pixel-to-micron calibration and numerical
aperture are not uniformly recorded in the acquisition metadata; the
characteristic object sizes at acquisition time
($\sim$10--14\,\textmu m WBCs $\to$ 60--120\,px,
ring parasites $\to$ 10--30\,px on 2048$^2$ frames) are consistent with a
$100{\times}$ oil-immersion objective, following standard thick-film
microscopy practice.

\noindent\textbf{Test set and evaluation sample.}
The held-out test set contains 410 thick-smear images
(134 Po, 125 Pm, 125 Pf, 26 Pv). For SGMCE evaluation we use an adaptive
per-species stratified sample: images are first ranked by detector density,
and the per-species budget is chosen so that fewer images are needed for
dense species (Pf) and more for sparse species (Pv, Pm). The Pv target of
95 detections is reached by including the entire available Pv test partition
(26 images); remaining budgets are allocated proportionally. The resulting
sample comprises 139 images and 737 detections at confidence threshold
$\hat{p} \geq 0.25$ (Pf: 233, WBC: 209, Pm: 105, Po: 97, Pv: 93).
All detections above the threshold -- including confidence-based
false-positive candidates -- are passed through the full pipeline.

\noindent\textbf{Configuration.}
YOLOv12n inference runs on an NVIDIA GPU; all subsequent stages are CPU and
API-bound. \emph{CropPreprocessor} targets a minimum crop dimension of 200\,px
with 50\,\% bounding-box padding for context. GPT-4o is queried at temperature
$0.05$ with up to 1{,}800 output tokens. The \emph{ValidationSuite} uses
\texttt{text-embedding-3-small} at cosine-similarity threshold $\theta=0.40$
with a shared embedding cache across all detections.

\subsection{Quantitative Results}
\label{sec:quantitative}

Table~\ref{tab:metrics} reports per-class mean validation scores.
Table~\ref{tab:metrics} (right panel)
compares sentence-level semantic scoring against a keyword-matching baseline
that tests only for verbatim KB strings, on the same 737 cached
explanations. Semantic scoring recovers most of the score lost to vocabulary
mismatch: GPT-4o paraphrases KB terms rather than quoting them verbatim
(\eg ``falciform'' for ``banana-shaped'', ``two chromatin dots'' for
``double chromatin'').

\begin{table}[tb]
\caption{Per-class validation scores and keyword-vs-semantic comparison
across 737 detections from 139 thick-smear images, using sentence-level
semantic scoring with $\theta = 0.40$. Left: overall KBC, CCF, DS per class.
Right: keyword matching vs.\ semantic scoring for KBC and DS on the same
GPT-4o explanations; the semantic rule with species-aware negation filtering
recovers most of the score lost to paraphrase. KBC and DS are not applicable
to WBC since the KB encodes no parasite concepts for this class.}
\label{tab:metrics}
\centering
\small
\begin{tabular}{l r r r r || r r r r}
\toprule
              &            &     &     &     & \multicolumn{2}{c}{KBC} & \multicolumn{2}{c}{DS} \\
\cmidrule(lr){6-7}\cmidrule(lr){8-9}
Species       & Detections & KBC & CCF & DS  & Keyword & Semantic & Keyword & Semantic \\
\midrule
\textit{P.\,falciparum} (Pf) & 233 & 0.81 & 1.00 & 0.96 & 0.02 & 0.81 & 0.57 & 0.96 \\
\textit{P.\,vivax}     (Pv)  &  93 & 1.00 & 0.90 & 1.00 & 0.01 & 1.00 & 0.34 & 1.00 \\
\textit{P.\,ovale}     (Po)  &  97 & 0.88 & 1.00 & 1.00 & 0.15 & 0.88 & 0.52 & 1.00 \\
\textit{P.\,malariae}  (Pm)  & 105 & 0.98 & 0.97 & 0.99 & 0.34 & 0.98 & 0.74 & 0.99 \\
WBC                          & 209 & n/a  & 0.94 & n/a  &      &      &      &      \\
\midrule
Parasite mean                & 528 & 0.91 & 0.97 & 0.99 & 0.13 & 0.91 & 0.54 & 0.99 \\
\bottomrule
\end{tabular}
\end{table}


\subsection{False Positive Analysis}
\label{sec:fp}

Because SGMCE passes every above-threshold detection to the explanation
pipeline, distractors and weakly supported detections are processed alongside
genuine parasites. We therefore report a confidence-based false-positive (FP)
flag as a diagnostic: any detection with $0.25 < \hat{p} \le 0.30$ (i.e., a
detection retained by the model, but only marginally above the detection
threshold) is flagged as a \emph{suspected} FP on the grounds that the detector
was least certain it corresponds to a real parasite. The flag is intentionally
conservative about the reverse direction where a low confidence does not prove
the object is a distractor, and some flagged detections are genuine parasites
the detector found visually difficult. The flagged subset provides a
short queue of candidates most worth a human reviewer's attention.
Across the 737 detections, 34 (4.6\,\%) were flagged
(Table~\ref{tab:fp}). Flag rates were highest for Pf (8.6\,\%) and
Pm (6.7\,\%), and lowest for Pv (1.1\,\%) and Po (1.0\,\%). For flagged
detections, the GPT-4o explainer tends to qualify its natural-language output
with caveats (the \texttt{caveats} field of the JSON schema), which a
human reviewer can use to decide whether to audit the detection manually.


\begin{table}[tb]
\caption{Confidence-based false-positive flag per class
($\hat{p} \leq 0.3$). The flag is a diagnostic aid, not a definitive FP
label: some low-confidence detections are genuine parasites that the detector
was simply less certain about.}
\label{tab:fp}
\centering
\small
\begin{tabular}{lrrr}
\toprule
Class & Low-conf. & High-conf. & Low-conf.\ rate \\
\midrule
Pf  &  20 & 213 & 8.6\,\% \\
Pv  &   1 &  92 & 1.1\,\% \\
Po  &   1 &  96 & 1.0\,\% \\
Pm  &   7 &  98 & 6.7\,\% \\
WBC &   5 & 204 & 2.4\,\% \\
\midrule
All &  34 & 703 & 4.6\,\% \\
\bottomrule
\end{tabular}
\end{table}

\subsection{Per-Rule CCF Diagnostic}
\label{sec:ccf_rules}

Aggregate CCF hides which individual correspondence rules are most active and
where the VLM's claims disagree with the CV measurements.
Table~\ref{tab:ccf_per_rule} reports, for each species and each activated
rule, the number of detections where the rule fired (the trigger keywords
appeared in the non-negated portion of the text) and the fraction for which
the CV measurement satisfied the rule condition. The ``pigment present'',
``single chromatin'', ``double chromatin'', ``round/compact'', and
``elongated shape'' rules fire on hundreds of detections and agree in
$\geq 99\,\%$ of cases across all four species, giving direct per-claim
evidence that GPT-4o's visual descriptions are grounded in the measured
morphology rather than being generic species labels.
The lowest per-rule agreement is ``irregular outline''$\to$\texttt{solidity}$<0.85$
for Pv, which fires on 10 detections and agrees on 0. Inspection shows
these are rings whose solidity exceeds 0.85 (CV measurement: compact) but
whose GPT-4o prose nonetheless describes them as ``slightly irregular''.
This is a disagreement the rule correctly catches, and is the primary driver
of Pv's CCF $=0.90$.

\begin{table}[tb]
\caption{Per-rule CCF diagnostic. For each species we show the six most-fired CCF rules, with the number of detections on which the rule activated and the
fraction that agreed with the CV measurement.
``--'' denotes rule-species pairs with zero activations.}
\label{tab:ccf_per_rule}
\centering
\small
\begin{tabular}{lrrrrc}
\toprule
Rule / Species & Pf & Pv & Po & Pm & ~~~~~~~~~~~~~~~~~~Agreement rate~~~~~~~~~~~~~~~~~~ \\
\midrule
pigment present           & 174 &  18 &  30 &  85 & 100\,\% (all) \\
round / compact           &  32 &  66 &  80 &  54 & 90.7\,\% (Pm); 100\,\% (Pf, Pv, Po) \\
single chromatin          & 102 &   2 &  12 &  13 & 100\,\% (all) \\
double chromatin          &  50 &   3 &   6 &  -- & 100\,\% (all) \\
elongated shape           &  10 &   2 &   1 &  -- & 100\,\% (all) \\
irregular outline         &  -- &  10 &  -- &   3 & 0\,\% (Pv); 66.7\,\% (Pm) \\
\bottomrule
\end{tabular}

\end{table}

\subsection{Qualitative Example}
\label{sec:qualitative}

\begin{figure}[tb]
\centering
\includegraphics[width=\textwidth]{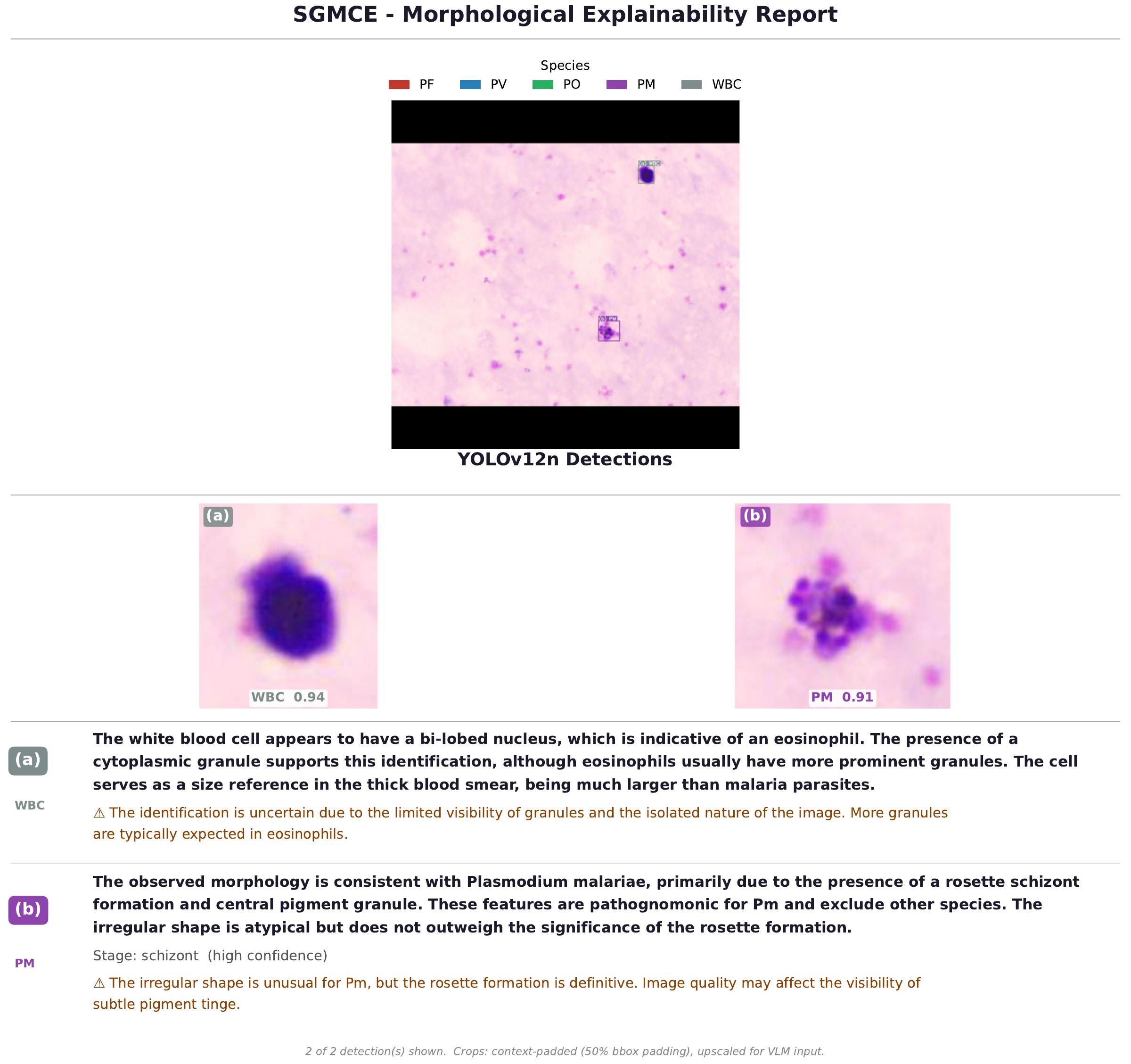}
\caption{Example SGMCE report on a thick-smear field with a Pm schizont and a co-detected WBC. 
For each detection, the report lists the detected class, the supporting morphological features, the free-form natural-language explanation, and the per-species differential reasoning.
}
\label{fig:report}
\end{figure}

Figure~\ref{fig:report} shows an example SGMCE report for a thick-smear field
in which the detector identified one Pm schizont ($\hat{p}=0.912$) and one
co-detected WBC ($\hat{p}=0.935$). For the Pm detection, SGMCE's supporting
features cite the rosette-pattern schizont and the central dark pigment
granule, both pathognomonic for Pm under the KB, and the differential block
rejects Pf (``no very small rings or double chromatin dots''), Pv (``no
amoeboid shape or golden-brown pigment''), and Po (``no early prominent
stippling; rosette not characteristic of Po''). The WBC detection, driven by
the dedicated WBC prompt, is described as a probable eosinophil
(bi-lobed nucleus, scant granules) and explicitly avoids parasite morphology
claims. The report contains no hand-curated text: the species call, the
supporting features, and the per-species rejections are all produced
automatically for each detection. This approach to explanation is arguably more generalizable to other use cases than the fixed text templates of MorphXAI.

\subsection{Discussion and Limitations}
\label{sec:discussion}

\paragraph{CCF analysis.}
CCF is $\geq 0.90$ for all parasite classes and 0.94 for WBCs
(Table~\ref{tab:metrics}). The per-rule diagnostic
(Table~\ref{tab:ccf_per_rule}) shows that the high-volume rules
(\texttt{pigment present}, chromatin-count rules, \texttt{round/compact})
fire on hundreds of detections and agree at $\geq 99\,\%$, so the aggregate
score is not carried by a few low-activation rules. The single visible
disagreement mode is the ``irregular outline'' $\to$ solidity rule for Pv
(Sect.~\ref{sec:ccf_rules}), which correctly flags a systematic over-
description by GPT-4o. Read together with the raw VLM text, CCF functions
as a per-claim auditor rather than a summary score.

\paragraph{KBC analysis.}
KBC under verbatim keyword matching is low (0.01--0.34 across species;
Table~\ref{tab:metrics}, right panel), not because the explanations omit thick-film
morphology but because GPT-4o paraphrases KB terms in natural clinical prose.
Sentence-level semantic scoring with species-aware negation filtering
recovers this substantially, raising parasite-class mean KBC from 0.13 to
0.91. The score is now dominated by the paraphrase coverage of the
explanation rather than by surface-form mismatch.

\paragraph{DS analysis.}
DS improves across all four species under semantic scoring
(Pf: $0.57\to 0.96$; Pv: $0.34\to 1.00$; Po: $0.52\to 1.00$;
Pm: $0.74\to 0.99$). The improvement is largest for species whose hallmarks
are vocabulary-rich but rarely quoted verbatim, notably the falciform
Pf gametocyte and the Pv amoeboid trophozoite.

\paragraph{Thick-film ambiguity and honest scope.}
The evaluation in this paper measures whether SGMCE's explanations are
consistent with the detector's predictions and with the thick-film KB. It does not claim to resolve cases that would be ambiguous even under expert review. The WHO bench aids themselves note that Pv, Po, and Pm are
often indistinguishable on thick film~\cite{who_benchaids}, and confirmatory
speciation is usually performed on a thin film. SGMCE inherits this limit
from the medium.

\paragraph{Negation handling.}
The CCF negation filter skips clauses such as ``unlike \textit{P.\,vivax},
no amoeboid cytoplasm'' before rule activation, and KBC/DS deliberately
exclude competing-species concepts from the score. These rules remove the
most visible inflation and deflation modes of embedding-based metrics, but
they do not replace a full natural-language inference model.

\paragraph{Life-stage inference.}
SGMCE exposes an \texttt{estimated\_life\_stage} field as a secondary output.
We do not evaluate per-stage accuracy quantitatively because the training
labels encode species only; reported life stages are provided as a
microscopist-facing aid but should not be interpreted as having been
validated against stage ground truth.

\paragraph{What this evaluation does not cover.}
No prospective expert-microscopist validation was conducted for this study;
the four automatic metrics are the primary evaluation. A human-subjects
study comparing unassisted microscopist review with SGMCE-assisted review
on real clinical slides is the natural next step. A separate extension is
applying the same pipeline to thin-film images, where intact RBCs enable
richer size-reference features.

\paragraph{Beyond malaria.}
The SGMCE template -- mask-guided crop preprocessing, handcrafted
morphological features, a thin-text domain KB, VLM reasoning, and an
automatic validation suite -- is not specific to \textit{Plasmodium}.
Any detection task where expert diagnosis is driven by structured
morphological features and where a concise KB can be written (\eg other
blood-film parasites, cytology, digital pathology ROI triage) fits the same
template with a different KB; the detector and the VLM remain unchanged.

\section{Conclusion}
\label{sec:conclusion}

We presented SGMCE, a post-hoc explanation framework for malaria species
identification in thick blood smears that generates clinically grounded
natural-language morphological explanations without model retraining and
without any additional morphological annotations. By coupling mask-guided
crop preprocessing, fourteen handcrafted adaptive CV morphological features grounded in clinical practice, and GPT-4o reasoning anchored in a WHO-derived thick-smear knowledge base,
SGMCE produces, for every detection, a species call with supporting features
and an explicit differential against each competing species. The KBC and DS metrics indicate that the explanations are grounded in clinically-derived rules, and the per-rule CCF diagnostic confirms that the VLM’s visual claims are grounded in the CV measurements rather than in generic species labels. Differential species reasoning, produced without
retraining or morphological supervision, directly supports microscopist
review, microscopist training, and the more general task of making object
detection in medical imaging auditable rather than opaque. More broadly, this method is generally applicable wherever explanations
of algorithm decisions are needed to increase user trust and
verifiability. Ultimately, by anchoring algorithmic decisions in verifiable domain
concepts, SGMCE offers a broadly applicable pathway to increase user
trust and support the responsible adoption of automated
decision-making systems.

\begin{credits}
\subsubsection{\ackname}

\subsubsection{\discintname}
The authors have no competing interests to declare that are relevant to the
content of this article.
\end{credits}

\bibliographystyle{splncs04}
\bibliography{sgmce_miua2026}

\end{document}